\documentclass[10pt]{article}
\usepackage[preprint]{tmlr}

\usepackage{amsmath,amsfonts,bm}

\def\eqref#1{equation~\ref{#1}}

\def\1{\bm{1}}

\DeclareMathAlphabet{\mathsfit}{\encodingdefault}{\sfdefault}{m}{sl}
\SetMathAlphabet{\mathsfit}{bold}{\encodingdefault}{\sfdefault}{bx}{n}

\usepackage{xcolor}
\usepackage[colorlinks=true, linkcolor=blue!60!black, citecolor=blue!60!black, urlcolor=blue!60!black]{hyperref}
\usepackage{url}
\usepackage{graphicx}
\usepackage{booktabs}
\usepackage{multirow}
\usepackage{amsmath,amssymb}
\usepackage{tikz}
\usetikzlibrary{arrows.meta,positioning}

\newcommand{\focus}{\textsc{focus}}
\newcommand{\fixed}{\textsc{fixed}}
\newcommand{\densewin}{\textsc{dense-win}}

\title{Anchors That Move Without Collapsing:\\Exclusive-Region Anchor Routing for Vision Transformers}

\author{\name Aadi Gupta \email aadi.24bcs10167@sst.scaler.com \\
      \addr Scaler School of Technology, Bengaluru, India}

\def\month{MM}
\def\year{YYYY}
\def\openreview{\url{https://openreview.net/forum?id=XXXX}}

\makeatletter
\newcommand{\codelink}{\if@accepted
  \url{https://github.com/ARRY7686/Partitioned-Residual-Anchor-Routing-Transformer}\else
  \url{https://anonymous.4open.science/r/XXXX} (anonymised for review)\fi}
\makeatother

\begin{document}

\maketitle

\begin{abstract}
Anchor tokens let a vision transformer exchange global information through a few summaries, but
nothing forces different anchors to summarise different things. On one shared backbone we show
that a Perceiver-style pathway and an AnchorFormer pathway collapse to a single summary, the
latter despite its mass normalisation (pooling-map similarity 0.99), while other soft pathways,
including Representative Attention, stay partly redundant (0.59--0.79). These results concern the
routing modules inside our block, not the full published models. An
exclusive partition of the patch grid cannot collapse by construction, and matched component by
component it is 2.4--3.6 points more accurate than Representative Attention on CIFAR-100. We
then propose \emph{focus anchors}, which place most anchors at the peaks of a learned importance
map while all anchors keep one exclusive Voronoi partition. Focus anchors beat a fixed partition
on a cluttered canvas (+1.5 points, 95\% CI [0.5, 2.6]) but not on full-HD traffic-sign
recognition. There, both variants match dense attention with local windows at about a fifth of
its FLOPs, and the fixed partition also trains 1.5$\times$ faster per step than dense attention
with FlashAttention, though both use more memory; a ViT on a downsampled input loses half its
mean average precision. Hard
exclusive assignment is the only region assignment we tested that is both stable and most
accurate: softmax and Sinkhorn relaxations diverge, and Gumbel-softmax trains but is less
accurate. We also report the limits: a ResNet-18 is more accurate on every task we trained it on
in this from-scratch regime, and
anchors train faster than FlashAttention only from about $1536^2$ pixels and never use less
memory.
\end{abstract}

\section{Introduction}

Global self-attention in a vision transformer (ViT)~\citep{dosovitskiy2021vit} costs $O(N^2)$ in
the number of patches $N$. A common remedy is to route global information through a small set of
$A\ll N$ \emph{anchor} (or latent, inducing, meta, representative) tokens: patches write into the
anchors, the anchors attend to each other, and the anchors write back
\citep{lee2019settransformer,jaegle2021perceiver,grainger2023paca,jiang2024lemevit,anchorformer2025,representative2026}.

The bottleneck only helps if the anchors carry \emph{different} information. In our first anchor
model, 32 anchors with learned queries converged to near-identical vectors (pairwise cosine
similarity 0.987--0.995 in every block) and each pooled from almost all 196 patches (attention
entropy 5.272 against a maximum of 5.278); Appendix~\ref{app:pilot} gives the setup. The model
had paid for 32 global summaries and received one. We call this \emph{anchor collapse}.

This paper asks how to prevent collapse while still letting anchors concentrate on image
content, and where such anchors help. Our contributions:
\begin{enumerate}
\item A controlled comparison of six anchor pathways on one backbone (Sec.~\ref{sec:collapse}),
including AnchorFormer and Representative Attention implemented from their papers with their
stabilisers. The Perceiver-style and AnchorFormer pathways collapse, AnchorFormer even with mass
normalisation; the other soft pathways stay partly redundant. An exclusive partition cannot
collapse, and a component-matched
ablation shows it is more accurate than Representative Attention, with or without the same
local convolution.
\item \emph{Focus anchors} (Sec.~\ref{sec:focus}): coverage anchors on a coarse fixed grid plus
anchors placed at the peaks of a learned importance map, all sharing one exclusive Voronoi
partition.
\item An evaluation on four tasks with paired confidence intervals (Secs.~\ref{sec:clutter}
and~\ref{sec:highres}). Focus anchors beat the fixed partition on Cluttered CIFAR-10 (and, not
significantly, on a Tiny ImageNet canvas), but not on Megapixel MNIST or full-HD traffic signs.
On full-HD images both anchor variants match dense attention at a fifth of its compute, while
downsampling halves accuracy.
\item A study of what makes moving anchors trainable (Sec.~\ref{sec:stability}): hard
exclusive assignment against softmax, Sinkhorn and Gumbel-softmax relaxations, with a
gradient-magnitude argument for the difference; reading from several anchors; placement
hyperparameters; and a reproducible instability with its fix.
\item A cost comparison against dense attention with FlashAttention~\citep{dao2022flashattention}
from $512^2$ to $2048^2$ (Sec.~\ref{sec:cost}), with the negative results that follow from it.
\end{enumerate}

\section{Related work}

\paragraph{Latent and anchor bottlenecks.}
Set Transformer inducing points~\citep{lee2019settransformer} and the
Perceiver~\citep{jaegle2021perceiver} cross-attend from a learned latent array to the input;
TokenLearner~\citep{ryoo2021tokenlearner} learns a few spatial attention maps that pool tokens;
Slot Attention~\citep{locatello2020slot} makes slots compete for inputs with a softmax over
slots. In ViTs, PaCa-ViT~\citep{grainger2023paca} attends from patches to learned clusters,
LeMeViT~\citep{jiang2024lemevit} uses learnable meta tokens, GPViT~\citep{yang2023gpvit}
propagates information through group tokens, and AnchorFormer~\citep{anchorformer2025} and
Representative Attention~\citep{representative2026} gather into anchors and distribute back,
Representative Attention with interaction among its representatives. These methods route with
soft, similarity-based weights, and several normalise the mass each representative receives to
keep all representatives in use. None assigns each patch to exactly one anchor. We test
AnchorFormer and Representative Attention, implemented from their papers, in
Sec.~\ref{sec:collapse}. VECA~\citep{song2026veca} also routes global information only through a
small set of learned cores, trained so that the number of cores can vary at inference, and
reports cores that organise semantically without supervision.

\paragraph{Regional tokens and partitions.}
RegionViT~\citep{chen2022regionvit} attaches one regional token to each fixed window of local
tokens and lets the regional tokens attend globally; our fixed partition is close to this design,
and we do not claim it as new. Swin~\citep{liu2021swin} and PVT~\citep{wang2022pvtv2} reduce
attention cost with windows and spatial reduction. Super tokens~\citep{huang2023supertoken} and
token clustering~\citep{zeng2022tcformer,zeng2024tcformer} adapt regions to the image but do not
keep a fixed number of exclusive regions with explicit positions. Bi-level routing
attention~\citep{zhu2023biformer} selects regions per query before token-level attention, and
DeBiFormer~\citep{baolong2024debiformer} adds deformable agent queries. We do not compare with
these methods or with VECA empirically.

\paragraph{Learned movement and selection.}
Deformable attention~\citep{xia2022dat} and progressive sampling~\citep{yue2021psvit} learn where
to sample keys or tokens; recurrent attention~\citep{mnih2014ram} and spatial
transformers~\citep{jaderberg2015stn} learn where to look. For high-resolution inputs under a
memory budget, attention sampling~\citep{katharopoulos2019attsampling} and differentiable or
iterative patch selection~\citep{cordonnier2021dps,bergner2023ips} process only selected patches.
Token pruning and merging~\citep{rao2021dynamicvit,bolya2023tome}, SparseViT~\citep{chen2023sparsevit},
mixture-of-depths~\citep{raposo2024mod}, A-ViT~\citep{yin2022avit} and ATS~\citep{fayyaz2022ats}
skip computation for some tokens. Focus anchors move where global summaries are formed, not
which patches exist. In multimodal language models, Lens~\citep{jiang2026lens} scores each visual
token's relevance to the question and softens low-relevance tokens with learned noise instead of
removing them; focus anchors also weight patches by a learned importance, but use it to decide
where global summaries are formed.

\section{Method}
\label{sec:method}

\subsection{Local--global--local block}

\begin{figure}[t]
\centering
\begin{tikzpicture}[
  box/.style={draw, rounded corners=2pt, minimum height=7mm, minimum width=19mm, font=\small, align=center},
  arr/.style={-{Stealth[length=2mm]}}, node distance=5mm]
\node[box] (local) {Local window\\attention};
\node[box, right=of local] (gather) {Gather\\(own region)};
\node[box, right=of gather, fill=gray!12] (global) {Anchor\\self-attention};
\node[box, right=of global] (scatter) {Scatter\\(own anchor)};
\node[box, right=of scatter] (ffn) {FFN};
\draw[arr] (local) -- (gather); \draw[arr] (gather) -- (global);
\draw[arr] (global) -- (scatter); \draw[arr] (scatter) -- (ffn);
\node[font=\small, below=1mm of gather] {$N\to A$};
\node[font=\small, below=1mm of global] {$A\times A$};
\node[font=\small, below=1mm of scatter] {$A\to N$};
\end{tikzpicture}
\caption{One block. Patches communicate locally inside windows and globally only through their
own anchor; the $N\times N$ attention of a ViT is replaced by $A\times A$ attention among anchors.}
\label{fig:block}
\end{figure}
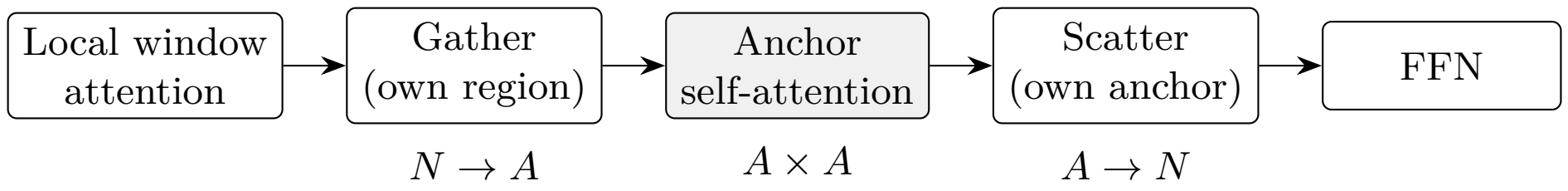

An image is split into a grid of $N$ patches with embeddings $x_i\in\mathbb{R}^{D}$ and centre
coordinates $c_i$, normalised by the long side of the grid. The model keeps $A$ anchors with
features $a_j\in\mathbb{R}^{d}$ ($d<D$) and positions $p_j$. Each of the $L$ blocks applies
(Fig.~\ref{fig:block}): (1) attention within non-overlapping $w\times w$ windows; (2)
\textbf{gather}: each anchor pools the patches of its own region $R_j$ with attention restricted
to $R_j$, and the pooled vector is added to $a_j$; (3) \textbf{global}: anchors attend to each
other, $a\leftarrow a+\mathrm{MHSA}(\mathrm{LN}(a))$, the only long-range interaction; (4)
\textbf{scatter}: each patch reads its own anchor $a_{r(i)}$ through a sigmoid gate,
$u_i=W_o\,\sigma(q_i^{\top}k_{r(i)}/\sqrt{d_h})\,v_{r(i)}$; (5) an FFN. Images are classified from
the mean of the final patch tokens, or their maximum where stated (Sec.~\ref{sec:highres}).

\subsection{Exclusive regions}

Every patch belongs to its nearest anchor, $r(i)=\arg\min_j\|c_i-p_j\|^2$, which partitions the
grid into Voronoi regions $R_j=\{i: r(i)=j\}$. The forward pass uses the hard assignment
$h_{ij}=[r(i)=j]$ with the gradient of the soft assignment
$s_{ij}=\mathrm{softmax}_j(-\|c_i-p_j\|^2/\tau)$, $\tau=(0.5/\sqrt{A})^2$
(straight-through~\citep{bengio2013straight}): $m_{ij}=h_{ij}+s_{ij}-\mathrm{sg}(s_{ij})$.
Anchor $j$ forms queries from $a_j+\phi(p_j)$, with $\phi$ a small MLP, and pools its region with
logits $\ell_{ji}$:
\begin{equation}
w_{ji}=\frac{m_{ij}\exp\!\big(\ell_{ji}-\max_{i'\in R_j}\ell_{ji'}\big)}
{\max\!\big(1,\ \sum_{i'} m_{i'j}\exp(\ell_{ji'}-\max_{i''\in R_j}\ell_{ji''})\big)} .
\label{eq:pool}
\end{equation}
A non-empty region always sums to at least 1, because its best patch contributes $\exp(0)=1$, so
the clamp at 1 acts only on \emph{empty} regions, which occur when two moving anchors meet. With
the usual $\epsilon=10^{-9}$ clamp, empty regions produced straight-through gradients of order
$10^{9}$ that made the first focus models diverge.

With anchors fixed on a grid, this block is the \emph{fixed partition} (\fixed). Because the
regions are disjoint, two anchors never pool the same patch, whatever the weights. This holds by
construction, so zero map overlap is not by itself evidence that anchors carry useful, different
information; we judge that by accuracy and by the spread of the pooled content (Eq.~\ref{eq:metrics}).

\subsection{Focus anchors}
\label{sec:focus}

A fixed partition gives the empty background as many anchors as the object. Focus anchors let
most anchors move to content while the partition stays exclusive. A linear head gives each patch
a score $z_i=w^{\top}\mathrm{LN}(x)_i$ and importance $\pi_i=\mathrm{softplus}(z_i)+10^{-3}$. In
every block:
\begin{itemize}
\item \textbf{Coverage anchors.} $g^2$ anchors stay on a fixed $g\times g$ grid,
$g=\max(1,\lfloor\sqrt{A/4}\rfloor)$, so every part of the image keeps a summary.
\item \textbf{Focus anchors.} The remaining $F=A-g^2$ anchors are placed one at a time by a
soft-argmax over the scores, $p_k=\sum_i \mathrm{softmax}_i(z_i+b_{ki})\,c_i$. After each
placement the scores around $p_k$ are suppressed,
$z_i\leftarrow z_i-\beta\exp(-\|c_i-p_k\|^2/2\sigma_s^2)$, so the next anchor goes elsewhere
($\beta=10$, $\sigma_s=1.5$ patches). The prior $b_{ki}=-\|c_i-e_k\|^2/2\sigma_h^2$ pulls each
focus anchor weakly towards a spread-out home position $e_k$ ($\sigma_h=0.35$) and breaks ties
early in training.
\item \textbf{Importance in pooling.} Gather logits are offset by $\log\pi_i$, so anchors pool
important patches more strongly; this also trains the importance head from the task loss.
\end{itemize}
All $A$ anchors then share the single Voronoi partition above: focus anchors can crowd onto
content, making their regions small, but never share a patch.

\section{Anchor collapse across anchor designs}
\label{sec:collapse}

\paragraph{Setup.} CIFAR-100~\citep{krizhevsky2009cifar} at $224^2$ (196 patches), 16 anchors,
$7\times7$ windows, 10 epochs, three seeds; every pathway sits in the same block, and only the
anchor pathway changes. \textsc{perceiver}: learned anchor queries, softmax over patches (the
pilot design). \textsc{competitive}: the same with a softmax over anchors, as in slot attention.
\textsc{anchorformer}: AnchorFormer's per-head learned anchors, softmax over anchors and
mass-normalised round trip $H=A\Delta^{-1}A^{\top}V$, its only described stabiliser.
\textsc{rpattention}: Representative Attention's slot-mass-normalised gather, latent
self-attention, query-driven distribution and $3\times3$ depth-wise convolution bypass.
\textsc{paca}: an MLP cluster assignment. \textsc{fixed}: the exclusive $4\times4$ partition. A
\textsc{dense} pathway replaces the anchors with global attention among all patches (local windows
kept). Neither AnchorFormer nor Representative Attention has released code; we implemented both
from their equations, with matched width and heads, as the global pathway of our block rather
than as the full published models (Appendix~\ref{app:reimpl}).

\paragraph{Metrics.} For one image and block, let $P_j\in\Delta^{N-1}$ be anchor $j$'s pooling
distribution and $g_j\in\mathbb{R}^{d}$ the summary it pools. We report
\begin{equation}
\mathrm{sim}=\frac{1}{A(A-1)}\sum_{j\neq k}\frac{\langle P_j,P_k\rangle}{\|P_j\|\,\|P_k\|},
\qquad
\mathrm{spread}=\frac{\sum_j\|g_j-\bar g\|^2}{\sum_j\|g_j\|^2},\quad \bar g=\frac{1}{A}\sum_j g_j ,
\label{eq:metrics}
\end{equation}
averaged over 1{,}024 test images and the 6 blocks. Similarity 1 means all anchors pool the same
patches; spread is the share of the summaries' energy that differs between anchors (0 = all
anchors carry the same vector). Unlike map similarity, spread is not fixed by the partition.

\begin{table}[t]
\centering
\caption{Anchor collapse on CIFAR-100 at $224^2$ (16 anchors, 10 epochs). Mean $\pm$ std over
seeds 42/43/44; similarity and spread averaged over blocks and seeds. Bottom: the $3\times3$
depth-wise convolution bypass of Representative Attention removed from it, or added to the
partition.}
\label{tab:collapse}
\begin{tabular}{lccc}
\toprule
Pathway & Test accuracy (\%) & Map similarity & Spread \\
\midrule
\textsc{perceiver} & 40.92 $\pm$ 0.30 & 0.998 & 0.00 \\
\textsc{anchorformer} & 42.52 $\pm$ 0.68 & 0.990 & 0.06 \\
\textsc{competitive} & 42.14 $\pm$ 0.67 & 0.792 & 0.22 \\
\textsc{paca} & 42.02 $\pm$ 0.07 & 0.592 & 0.18 \\
\textsc{rpattention} (with its conv) & 47.54 $\pm$ 0.73 & 0.683 & 0.27 \\
\textsc{dense} (no anchors) & 40.93 $\pm$ 0.95 & -- & -- \\
\fixed{} & 46.31 $\pm$ 0.56 & 0.000 & 0.60 \\
\midrule
\textsc{rpattention} without conv & 42.68 $\pm$ 0.86 & 0.687 & 0.28 \\
\fixed{} with conv & \textbf{49.93 $\pm$ 0.71} & 0.000 & 0.59 \\
\bottomrule
\end{tabular}
\end{table}

\begin{figure}[t]
\centering
\includegraphics[width=\textwidth]{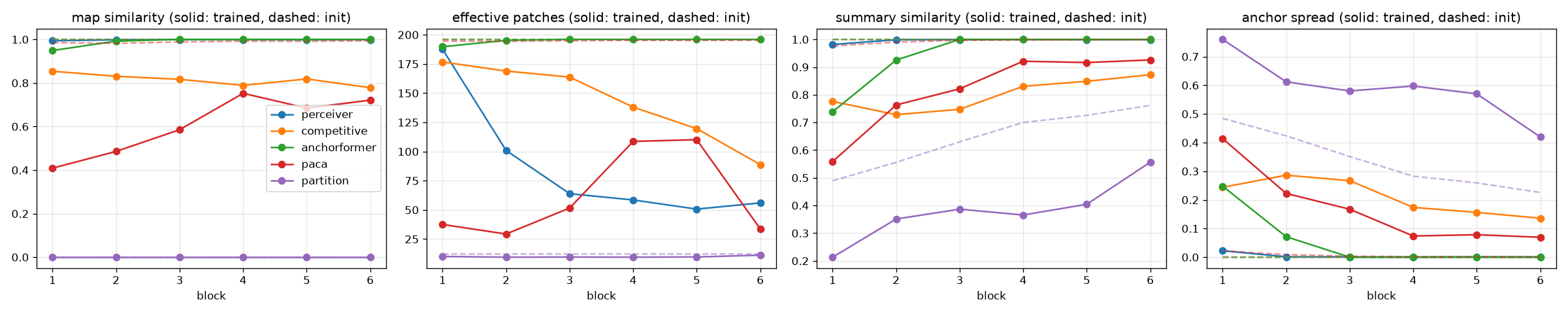}
\caption{Collapse metrics per block (seed 42), trained (solid) and at initialisation (dashed),
for the pathways of the original comparison. The \textsc{perceiver} and \textsc{anchorformer}
pathways converge to the same summary; the fixed partition keeps zero map similarity and the
highest spread.}
\label{fig:collapse}
\end{figure}

\paragraph{Results.} Table~\ref{tab:collapse} and Fig.~\ref{fig:collapse}. The
\textsc{perceiver} and \textsc{anchorformer} pathways collapse almost completely: map similarity
0.990--0.998 and near-zero spread, AnchorFormer despite its mass normalisation. The
\textsc{competitive}, \textsc{paca} and \textsc{rpattention} pathways stay partly diverse
(0.59--0.79), but their summaries grow more similar with depth; Representative Attention's
stabilisers prevent full collapse, yet its summaries are far less distinct than the partition's
(spread 0.27 against 0.60).

Our data do not single out one cause. AnchorFormer and Representative Attention route almost
identically: both score each key against learned prototypes shared by all images, take a softmax
over them and normalise each prototype's mass. They differ after gathering: Representative
Attention's representatives attend to each other, and patches read them by attention over their
keys, whereas AnchorFormer reads back through the same assignment in reverse
($H=A\Delta^{-1}A^{\top}V$). A read-back by attention alone does not prevent collapse, however:
the \textsc{perceiver} pathway also reads back by attention, and it collapses. We have not
isolated which combination of competition between anchors, interaction among them and the form
of the read-back keeps anchors distinct; the exclusive partition avoids the question by
construction. These results describe the routing modules in our block, trained from scratch on
CIFAR-100. They do not show that the published AnchorFormer or Representative Attention models
collapse or underperform as designed; both papers report results on ImageNet and dense-prediction
benchmarks, which we did not reproduce.

Representative Attention is more accurate than the plain partition (+1.23 points, 95\% CI
[+0.02, +2.45]), but it also has a $3\times3$ depth-wise convolution that the other pathways lack.
Removing it costs Representative Attention 4.9 points ([+1.0, +8.7]); adding the same
convolution to the partition gains 3.6 ([+1.3, +5.9]). Matched in this component, the exclusive
partition is ahead by 2.39 points with the convolution ([+0.86, +3.91]) and by 3.63 without it
([+0.14, +7.11]), and ahead of the dense pathway by 5.4. In our setting, exclusive routing gives
the most distinct summaries and the best accuracy of the anchor designs we tested; a cheap local
convolution helps every pathway.

\section{Focus anchors on cluttered canvases}
\label{sec:clutter}

\paragraph{Tasks.} In \emph{Cluttered CIFAR-10}, a $32^2$ CIFAR-10 image is upsampled to $64^2$
and pasted at a random position on a black $256^2$ canvas among 8 distractor fragments (random
$16^2$ crops of other training images, upsampled to $32^2$); the object covers 6.25\% of the
canvas. In the \emph{Tiny ImageNet canvas}, a $64^2$ Tiny ImageNet~\citep{le2015tinyimagenet}
image (200 classes) is pasted without resizing among 8 native $32^2$ crops, so downsampling the
canvas also removes real detail. Seeds change initialisation, data order and canvas layouts; test
canvases use a fixed layout seed.

\paragraph{Models.} Besides \focus{} and \fixed{}, we train a dense ViT of the same width and
depth (global attention in every block, no windows) on the full canvas and on area-downsampled
$128^2$ and $64^2$ inputs, \densewin{} (our blocks with the anchor pathway replaced by global
attention among all patches, local windows kept), and, from outside our family,
PVTv2-B0~\citep{wang2022pvtv2} and ResNet-18~\citep{he2016resnet} (no pretraining). Training
recipes are in Appendix~\ref{app:recipes}. Experiments in this section have no validation split;
every setting was fixed in advance except the 30-epoch learning rate. At $3\cdot10^{-4}$ about
one in six 30-epoch runs diverged, and we chose $1.25\cdot10^{-4}$, the DeiT scaling rule~\citep{touvron2021deit}, after comparing the test
accuracy of one run at each rate at epoch 15. That rate was used for every model, and ResNet-18's
better of two rates is reported, so these numbers are slightly optimistic for all models alike.

\begin{table}[t]
\centering
\caption{Cluttered CIFAR-10 and Tiny ImageNet canvas: test accuracy (\%), mean $\pm$ std over 3
seeds. GFLOPs per image include attention. Bottom: paired gaps, mean [95\% CI].}
\label{tab:main}
\begin{tabular}{lrrccc}
\toprule
& & & \multicolumn{2}{c}{Cluttered CIFAR-10} & Tiny IN \\
Model & Params & GFLOPs & 10 ep. & 30 ep. & 15 ep. \\
\midrule
\focus-16 & 11.4M & 5.49 & 49.84 $\pm$ 0.26 & \textbf{66.82 $\pm$ 0.66} & 17.68 $\pm$ 0.33 \\
\focus-4 & 11.4M & 5.45 & 47.52 $\pm$ 1.11 & 64.67 $\pm$ 0.26 & -- \\
\fixed-16 & 11.4M & 5.49 & 46.71 $\pm$ 1.72 & 65.29 $\pm$ 0.94 & 16.73 $\pm$ 0.46 \\
\densewin & 12.8M & 7.25 & -- & 65.10 $\pm$ 0.45 & \textbf{20.50 $\pm$ 0.29} \\
ViT, $256^2$ & 9.3M & 5.28 & 46.74 $\pm$ 0.42 & 61.99 $\pm$ 0.54 & 18.25 $\pm$ 0.33 \\
ViT, $128^2$ & 9.2M & 1.21 & 43.20 $\pm$ 0.22 & 55.05 $\pm$ 0.26 & 12.49 $\pm$ 0.28 \\
ViT, $64^2$, patch 8 & 9.0M & 1.18 & 42.39 $\pm$ 0.41 & -- & 13.01 $\pm$ 0.39 \\
\midrule
PVTv2-B0 & 3.4M & 1.51 & 71.83 $\pm$ 0.30 & 75.07 $\pm$ 0.25 & -- \\
ResNet-18 & 11.2M & 4.74 & -- & 86.11 $\pm$ 0.26 & 37.07 $\pm$ 0.28 \\
\bottomrule
\end{tabular}

\vspace{2mm}
\begin{tabular}{lcc}
\toprule
Paired gap & Cluttered, 30 ep. & Tiny IN, 15 ep. \\
\midrule
\focus-16 $-$ \fixed-16 & +1.54 [+0.53, +2.55] & +0.95 [$-$0.93, +2.83] \\
\focus-16 $-$ \densewin & +1.72 [$-$0.80, +4.25] & $-$2.82 [$-$3.99, $-$1.66] \\
\fixed-16 $-$ \densewin & +0.19 [$-$2.71, +3.08] & $-$3.77 [$-$4.65, $-$2.90] \\
\focus-16 $-$ ViT $128^2$ & +11.77 [+9.73, +13.82] & +5.19 [+4.43, +5.95] \\
\bottomrule
\end{tabular}
\end{table}

\begin{figure}[t]
\centering
\includegraphics[width=\textwidth]{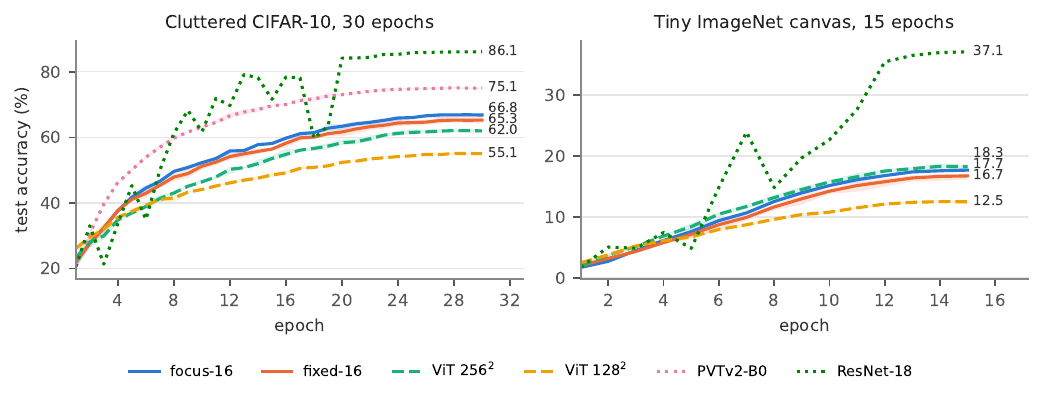}
\caption{Test accuracy during training, mean over 3 seeds (band: min--max; omitted for ResNet-18,
whose per-epoch test accuracy fluctuates strongly). Anchor models solid, baselines dashed or
dotted.}
\label{fig:curves}
\end{figure}

\paragraph{Focus against fixed.} Table~\ref{tab:main} and Fig.~\ref{fig:curves}. At 10 epochs focus anchors beat the fixed
partition in all six paired runs (4 and 16 anchors; per-seed values in
Appendix~\ref{app:perseed}). At 30 epochs they win in all three seeds (+1.1 to +1.8 points), and
the interval excludes zero. On the Tiny ImageNet canvas focus is ahead in all three seeds, but the
interval includes zero.

\paragraph{Against dense attention.} Dense attention with the same local windows (\densewin)
closes the anchors' lead over the plain dense ViT on Cluttered CIFAR-10: the fixed partition is
level with it (+0.19), and focus anchors are ahead in every seed without a significant gap
(+1.72 [$-$0.80, +4.25]), at 5.49 against 7.25 GFLOPs. On the Tiny ImageNet canvas \densewin{} is
the best transformer, 2.8 points ahead of focus anchors. So the anchors' earlier advantage over
the dense ViT was largely locality, and global attention among all patches is better when the
whole object carries fine detail.

\paragraph{Against downsampling and established models.} Focus anchors beat a ViT on a
$128^2$ or $64^2$ input by 4.7--11.8 points on both tasks. PVTv2-B0 and ResNet-18 are far more
accurate (75.1\% and 86.1\% against 66.8\% on Cluttered CIFAR-10); trained from scratch on
50k--100k images for 15--30 epochs, the convolutional bias of these models matters more than how
a plain transformer routes global information. Our claims are about anchor routing within
transformers of this kind.

\begin{figure}[t]
\centering
\includegraphics[width=\textwidth]{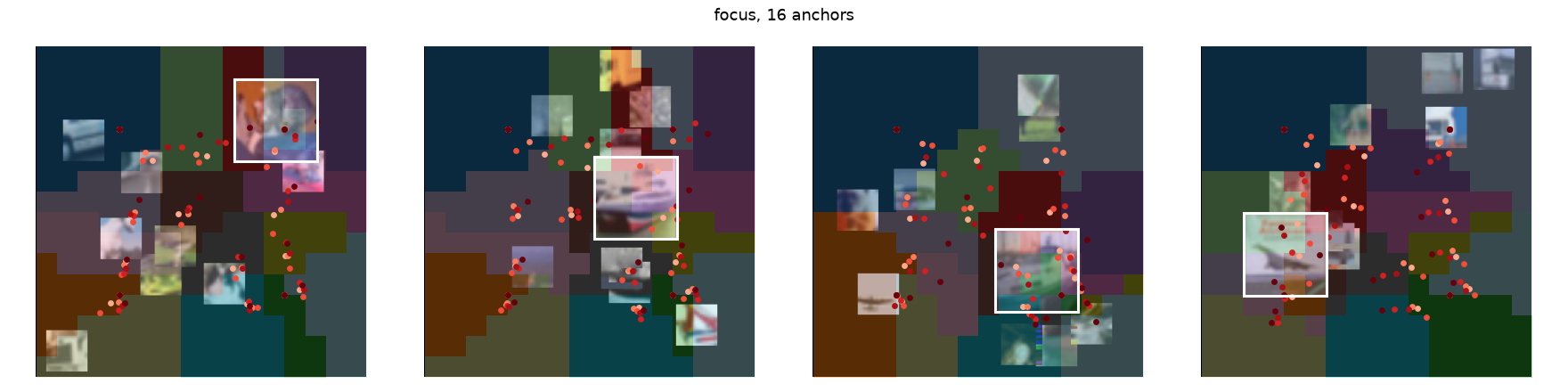}
\caption{Final-block regions (colours) and anchor positions in each block (dots, darker = later)
of \focus{}-16 on four test canvases; the white box is the object.}
\label{fig:anchors}
\end{figure}

\paragraph{Where the anchors go.} Fig.~\ref{fig:anchors}: focus anchors gather on \emph{all}
image content, the object and the distractors, and leave the empty background to a few large
coverage regions. The importance map learns where there is content, not where the class object
is, which is what we expect when the distractors are real image fragments. Conditioning the
importance on a query, as question-conditioned relevance scoring does in multimodal
models~\citep{jiang2026lens}, might steer anchors to the object; we have not tested this. With 4 anchors, the
share whose region contains the object rises from 0.44 in the first block to 0.57 in the last,
against 0.39 for the fixed partition. Moving anchors that take one importance-weighted Lloyd step
inside their own region per block find the object in their importance map but move only about one
patch, because a Lloyd step cannot leave its region: placement must be global, then partitioned.

\section{High-resolution images}
\label{sec:highres}

\paragraph{Tasks.} \emph{Megapixel MNIST}~\citep{katharopoulos2019attsampling}: five native $28^2$
MNIST~\citep{lecun1998mnist} digits, three of the target class and two of another, among 12
stroke-like noise patches on a $512^2$ canvas; the label is the majority digit. Training layouts
are drawn afresh every epoch (10{,}000 per epoch); validation and test use 1{,}000 fixed layouts
each, with digits from held-out MNIST training images and from the MNIST test set.
\emph{DFG traffic signs}~\citep{tabernik2020dfg}: $1920\times1080$ street scenes; we recognise
which of the 20 most frequent sign types are present (multi-label, mean average precision over
classes, chance 3.1). The official split is kept (4{,}480 training, 774 validation from blocks of
consecutive frames, 1{,}703 test frames); the median sign is 134 pixels wide and 10\% are under
45. Both tasks have validation splits and settings fixed in advance (AdamW, learning rate
$2\cdot10^{-4}$, batch 32, gradient clipping and warmup; ResNet-18 at $10^{-3}$); 30 and 20
epochs. Anchor models use 64 anchors (16 coverage, 48 focus) on rectangular patch grids, with
positions in units of the long side.

\paragraph{Pooling.} With average pooling over the final tokens, every model stayed near chance
on DFG (8--12 mAP), including an ImageNet-pretrained ResNet-18 (8.7). We checked the pipeline
(Appendix~\ref{app:dfg}): the metric matches a reference implementation, boxes align with signs,
and a pretrained ResNet-18 memorises 64 training frames to 100 mAP. With \emph{max} pooling the
same pretrained ResNet-18 reaches 47.8 mAP in 6 epochs: averaging over a full-HD frame washes out
signs that cover under 1\% of it. All DFG results below use max-pooled heads for every model.

\begin{table}[t]
\centering
\caption{High-resolution tasks, mean $\pm$ std over 3 seeds. Megapixel MNIST: test accuracy
(\%); focus anchors over seeds 43 and 44 (seed 42: Sec.~\ref{sec:stability}). DFG: mAP with
max-pooled heads on the official test split and on our validation split, which holds out blocks
of consecutive frames; GFLOPs per DFG image.}
\label{tab:highres}
\begin{tabular}{lcccr}
\toprule
& Megapixel MNIST & \multicolumn{2}{c}{DFG, $1920\times1080$} & \\
Model & $512^2$, test & test & validation & GFLOPs \\
\midrule
\focus-64 & 91.15 $\pm$ 0.92 & \textbf{30.28 $\pm$ 0.88} & 38.89 $\pm$ 1.52 & 187 \\
\fixed-64 & \textbf{91.87 $\pm$ 0.35} & 29.62 $\pm$ 0.20 & \textbf{40.00 $\pm$ 1.30} & 187 \\
\densewin & 91.40 $\pm$ 0.20 & \textbf{30.28 $\pm$ 0.85} & 38.92 $\pm$ 1.65 & 912 \\
Dense ViT, full resolution & 86.10 $\pm$ 0.96 & 28.60 $\pm$ 0.64 & 39.54 $\pm$ 0.88 & 846 \\
ViT, $256^2$ / $256$-wide & 93.37 $\pm$ 2.40 & -- & -- & -- \\
ViT, $128^2$ / $480$-wide & 87.50 $\pm$ 1.90 & 15.80 $\pm$ 0.67 & 18.66 $\pm$ 0.05 & 13 \\
\midrule
PVTv2-B0 & 97.43 $\pm$ 0.21 & 19.50 $\pm$ 1.75 & 22.29 $\pm$ 3.29 & 195 \\
ResNet-18 & 99.30 $\pm$ 0.26 & 36.51 $\pm$ 1.65 & 44.48 $\pm$ 0.54 & 160 \\
\bottomrule
\end{tabular}
\end{table}

\paragraph{Results.} Table~\ref{tab:highres}. On DFG, downsampling halves accuracy: a ViT on a
480-pixel-wide input reaches 15.8 mAP against 28.6--30.3 for the anchor models and both dense
models at full resolution (fixed $-$ downsampled ViT +13.8 [+12.6, +15.1]). Both anchor models match dense attention at
about a fifth of its FLOPs: focus $-$ \densewin{} +0.00 [$-$4.2, +4.2], fixed $-$ dense ViT +1.0
[$-$1.1, +3.1]. Focus and fixed anchors are not distinguishable (+0.67 [$-$1.90, +3.23]). The
official split interleaves frames from the same video sequences, so test images can be
near-duplicates of training images; on our validation split, which holds out whole blocks of
frames, every conclusion is the same (downsampled ViT 18.7 against 38.9--40.0 at full resolution,
anchors level with dense attention, ResNet-18 best), and the near-duplicates are shared by all
models alike. On
Megapixel MNIST the anchor models and \densewin{} are level (fixed $-$ \densewin{} +0.47
[$-$0.25, +1.18]) and 5.3--5.8 points above the dense ViT without windows. At $512^2$ the digits
remain legible at half resolution, and the $256^2$ ViT is not significantly different from the
anchors (+1.50 [$-$3.6, +6.6]); only at $128^2$ (digits of 7 pixels) does downsampling cost
accuracy (fixed $-$ ViT $128^2$ +4.37 [+0.47, +8.27]). ResNet-18 is best on both tasks; PVTv2 is
strong on Megapixel MNIST but weak on DFG, where its early spatial reduction may lose small signs.

Taken with Sec.~\ref{sec:clutter}: moving the anchors paid off only on Cluttered CIFAR-10, a
single object among real image fragments, and not on Megapixel MNIST's several digits among
strokes; at high resolution the value of anchors is keeping full resolution at a fraction of the
cost of dense attention, not better accuracy than it.

\section{Cost}
\label{sec:cost}

\begin{table}[t]
\centering
\caption{Training step at batch 8 in ms (peak training memory in GB), untrained models, one A100,
bf16, measured alone. Dense ViT with the fused FlashAttention kernel or with plain attention.}
\label{tab:cost}
\begin{tabular}{lcccc}
\toprule
Resolution & Dense ViT, FlashAttention & Dense ViT, plain & \fixed-64 & \focus-64 \\
\midrule
$512^2$ & 11.5 (0.9) & 44.8 (2.6) & 44.7 (1.4) & 196.9 (1.5) \\
$1024^2$ & 58.9 (3.0) & 548.6 (30.0) & 68.5 (4.8) & 196.7 (5.3) \\
$1536^2$ & 198.4 (6.3) & out of memory & 134.9 (10.5) & 296.2 (11.5) \\
$2048^2$ & 533.6 (11.1) & out of memory & 232.1 (18.5) & 288.2 (20.2) \\
$1920\times1152$ & 187.9 (6.4) & out of memory & 128.0 (10.4) & 204.2 (11.3) \\
\bottomrule
\end{tabular}
\end{table}

Table~\ref{tab:cost}. Against plain attention, whose memory grows quadratically, the anchor models
are much cheaper and the dense ViT runs out of memory from $1536^2$. Against
FlashAttention~\citep{dao2022flashattention}, the fair comparison, the picture is narrower. The
fixed partition trains faster from about $1536^2$: 1.5$\times$ at full HD and 2.3$\times$ at
$2048^2$. Focus anchors only overtake it at $2048^2$, because placing 48 anchors one at a time
adds 56--161 ms per step that does not shrink with resolution. Both anchor models always use
\emph{more} training memory than FlashAttention (18.5--20.2 against 11.1 GB at $2048^2$): our
masked gather and scatter are not fused into memory-efficient kernels. At $256^2$
(Appendix~\ref{app:cost}) dense attention is cheap and the anchors are 1.9--3.1$\times$ slower.
Established efficient transformers are faster still: in an earlier inference measurement at
$2048^2$, PVTv2-B1 and EfficientViT-B2~\citep{cai2023efficientvit} were 2$\times$ and 8$\times$
faster than the fixed partition.

\section{What makes moving anchors trainable}
\label{sec:stability}

\begin{table}[t]
\centering
\caption{Routing variants of \focus-16 on Cluttered CIFAR-10, 10 epochs, gradient clipping 1.0
and 2-epoch warmup, seeds 42/43/44. Paired gaps against the reference, mean [95\% CI].}
\label{tab:ablation}
\begin{tabular}{lcl}
\toprule
Variant & Test accuracy (\%) & Note \\
\midrule
Reference: hard assignment, own anchor & 53.78 $\pm$ 0.09 & \\
Soft (softmax) assignment & 10.00 / 10.00 / 10.00 & diverged in all seeds \\
Soft + Sinkhorn balancing & 9.98 / 10.00 / 12.44 & peaked at 26--39\%, then diverged \\
Soft Gumbel-softmax, annealed & 22.08 / 49.53 / 51.30 & seed 42 diverged at epoch 9 \\
Read from 2 nearest anchors & 53.76 $\pm$ 0.19 & $-$0.03 [$-$0.35, +0.30] \\
Read from 3 nearest anchors & 54.61 $\pm$ 0.55 & +0.83 [$-$0.55, +2.20] \\
No suppression & 53.57 $\pm$ 0.43 & $-$0.21 [$-$1.51, +1.08] \\
\bottomrule
\end{tabular}
\end{table}

\paragraph{Hard against soft assignment.} Table~\ref{tab:ablation}. Replacing the hard
straight-through assignment by its softmax relaxation diverged in every seed, with gradient norms
up to $10^{14}$. Balancing the soft assignment with three log-domain Sinkhorn
iterations~\citep{cuturi2013sinkhorn}, so that every anchor receives the same mass, delays but
does not prevent divergence. Gumbel-softmax~\citep{jang2017gumbel} with a temperature annealed
from 1.0 to 0.1 is the only soft variant that trains: in two seeds it ends 2.5 and 4.3 points
below the hard reference, and in the third it trained normally to epoch 8 and diverged as the
temperature approached 0.1. The closer the soft assignment gets to hard, the larger its gradients;
the straight-through hard assignment has the hard forward pass without that instability. Hard
exclusive assignment is the only region assignment we tested that is both stable and most
accurate.

\paragraph{Why the straight-through assignment stays stable.} A gradient-magnitude argument
explains part of this. Under the straight-through assignment the forward pass uses only the
hard regions, and the gradient reaches an anchor position $p_k$ only through the soft
assignment at the fixed temperature $\tau$:
\begin{equation}
\frac{\partial s_{ij}}{\partial p_k}=\frac{2}{\tau}\,s_{ij}\,(\delta_{jk}-s_{ik})\,(c_i-p_k),
\qquad
\Big\|\frac{\partial s_{ij}}{\partial p_k}\Big\|\le\frac{2}{\tau}\cdot\frac14\cdot\sqrt2
=\frac{\sqrt2}{2\tau}=2\sqrt2\,A ,
\label{eq:stbound}
\end{equation}
since $|s_{ij}(\delta_{jk}-s_{ik})|\le\frac14$, $\|c_i-p_k\|\le\sqrt2$ in normalised
coordinates, and $\tau=(0.5/\sqrt A)^2$. The pooling weights of Eq.~\ref{eq:pool} divide by a
denominator of at least 1 and use $e^{\ell-\max\ell}\le1$, so each derivative
$\partial w/\partial m$ is at most 2 in magnitude, and the factor $\partial\mathcal{L}/\partial m$
that multiplies this Jacobian is bounded by a small multiple of the loss gradient on the pooled
values. Neither factor depends
on how sharp the partition is or on how close two anchors are. A relaxation that uses the soft
assignment in the forward pass at temperature $T$ has the same Jacobian with $\tau$ replaced by
$\tau T$, so its bound grows as $1/T$, and it must sharpen ($T\to0$) to approach the hard
partition that performs best. This matches the Gumbel-softmax runs, which trained normally
while $T$ was large and diverged in one seed as $T$ approached 0.1, a tenfold larger bound. The
argument does not explain why the plain softmax relaxation also diverges at $T=1$; there the
forward pass mixes neighbouring regions, and focus anchors pulled towards the same content keep
many patches near region boundaries, where $s(1-s)$ is largest, but we have not isolated this
mechanism.

\paragraph{Reading from several anchors.} If each patch reads from its $k$ nearest anchors,
weighted by the distance softmax, while gathering stays exclusive, $k=2$ changes nothing and
$k=3$ is ahead in all three seeds by +0.83 points without a significant gap. Exclusivity matters
on the gather side; the scatter side can be relaxed.

\paragraph{Placement.} Suppression does not change accuracy once training is stabilised, but
without clipping one of three runs without suppression diverged; the home prior adds 2.6 points
in the original recipe; the coverage grid makes no measurable difference. Placement is robust
within about a factor of two of $\beta$, $\sigma_s$ and $\sigma_h$, and breaks when suppression
is much stronger ($\beta=30$, $-$4.4 points) or wider ($\sigma_s=3$ patches, training stalls near
20\%) (Appendix~\ref{app:sensitivity}).

\paragraph{A reproducible instability.} On Megapixel MNIST at learning rate $2\cdot10^{-4}$, focus
anchors trained cleanly for seeds 43 and 44 (90.5\% and 91.8\%), but seed 42 became unstable at
epoch 13: the median gradient norm rose from about 5 to over 1{,}000, clipping kept the run from
diverging, and it ended at 79.0\%. Training is deterministic for a seed and schedule; a rerun
reproduced the run exactly. The largest gradients appeared first in the patch embedding and the
first block's local attention, with the importance head among them. At learning rate
$10^{-4}$ the same seed trained without a single gradient spike (88.7\%); with 16 instead of 64
anchors at $2\cdot10^{-4}$ it was worse (64.4\%). In this case the instability depended on the
step size rather than on the number of focus anchors; we recommend the lower rate for focus
anchors at this scale, but have tested it on this one seed only.

\section{Limitations}

All models are trained from scratch for 10--30 epochs on at most 100k images; we have no
pretraining or ImageNet-scale results, and a ResNet-18 is more accurate on every task we trained
it on. Three seeds
give wide intervals; the sensitivity sweep and several skipping rows use one or two seeds.
Sections~\ref{sec:collapse}--\ref{sec:clutter} have no validation split and one learning rate
was chosen on test accuracy; Sec.~\ref{sec:highres} uses validation splits and settings fixed in
advance. AnchorFormer and Representative Attention are implemented from their papers, without
official code, and used as the global pathway of our block rather than as the full published
models; the PaCa-style pathway is simplified. Our conclusion that hard assignment is the most
stable and accurate holds for the relaxations we tested (softmax, Sinkhorn, Gumbel-softmax); we
did not test diversity or orthogonality regularisers on soft anchors, nor routing methods such
as BiFormer, DeBiFormer or VECA. DFG's official split interleaves frames from the
same video sequences (our block-held-out validation split gives the same conclusions), and its
image-level labels make it a weakly supervised task; we needed max
pooling for any model to learn it. The importance map marks content in general rather than the
class object, and it does not choose patches for computation better than random
(Appendix~\ref{app:skipping}). Our implementation is not fused: below about $1536^2$ it is slower
than FlashAttention, and it never uses less memory.

\section{Conclusion}

Anchor bottlenecks only pay off if the anchors stay different. Perceiver-style and AnchorFormer
anchors collapse to a single summary, the latter even with mass normalisation; other soft
pathways, Representative Attention included, stay partly redundant, and we could not tie the
difference to a single design choice. An exclusive partition prevents collapse by construction and, matched
component by component, routes most accurately among the designs we tested. Focus anchors keep
that guarantee while moving to image content; they beat a fixed partition on a cluttered CIFAR-10
canvas, though not on our high-resolution tasks, and on full-HD images both keep full resolution
at a fifth of the cost of dense attention, where downsampling halves accuracy. Hard exclusive assignment is what makes moving
anchors trainable. The gains are within transformers trained from scratch; next steps are fused
kernels and parallel placement, pretrained backbones, tasks that need long-range relations
between distant small objects, and video, where most tokens are uninformative.

\subsubsection*{Broader Impact Statement}
This work studies an architectural component of vision transformers on public benchmarks
(CIFAR, Tiny ImageNet, MNIST, DFG traffic signs). We do not foresee direct negative impacts
beyond those of image recognition models in general. Traffic-sign recognition is safety-relevant;
our models are research prototypes far from the accuracy required for deployment.

\paragraph{Code availability.} Code, per-run logs and result files for every experiment:
\codelink.

\bibliography{refs}

\begin{thebibliography}{44}
\providecommand{\natexlab}[1]{#1}
\providecommand{\url}[1]{\texttt{#1}}
\expandafter\ifx\csname urlstyle\endcsname\relax
  \providecommand{\doi}[1]{doi: #1}\else
  \providecommand{\doi}{doi: \begingroup \urlstyle{rm}\Url}\fi

\bibitem[BaoLong et~al.(2024)BaoLong, Zhang, Shi, Hirakawa, Yamashita, Matsui, and Fujiyoshi]{baolong2024debiformer}
NguyenHuu BaoLong, Chenyu Zhang, Yuzhi Shi, Tsubasa Hirakawa, Takayoshi Yamashita, Tohgoroh Matsui, and Hironobu Fujiyoshi.
\newblock {DeBiFormer}: Vision transformer with deformable agent bi-level routing attention.
\newblock In \emph{Asian Conference on Computer Vision (ACCV)}, pp.\  445--462, 2024.

\bibitem[Bengio et~al.(2013)Bengio, L{\'e}onard, and Courville]{bengio2013straight}
Yoshua Bengio, Nicholas L{\'e}onard, and Aaron Courville.
\newblock Estimating or propagating gradients through stochastic neurons for conditional computation.
\newblock arXiv:1308.3432, 2013.

\bibitem[Bergner et~al.(2023)Bergner, Lippert, and Mahendran]{bergner2023ips}
Benjamin Bergner, Christoph Lippert, and Aravindh Mahendran.
\newblock Iterative patch selection for high-resolution image recognition.
\newblock In \emph{International Conference on Learning Representations (ICLR)}, 2023.

\bibitem[Bolya et~al.(2023)Bolya, Fu, Dai, Zhang, Feichtenhofer, and Hoffman]{bolya2023tome}
Daniel Bolya, Cheng-Yang Fu, Xiaoliang Dai, Peizhao Zhang, Christoph Feichtenhofer, and Judy Hoffman.
\newblock Token merging: Your {ViT} but faster.
\newblock In \emph{International Conference on Learning Representations (ICLR)}, 2023.

\bibitem[Cai et~al.(2023)Cai, Li, Hu, Gan, and Han]{cai2023efficientvit}
Han Cai, Junyan Li, Muyan Hu, Chuang Gan, and Song Han.
\newblock {EfficientViT}: Lightweight multi-scale attention for high-resolution dense prediction.
\newblock In \emph{IEEE/CVF International Conference on Computer Vision (ICCV)}, 2023.

\bibitem[Chen et~al.(2022)Chen, Panda, and Fan]{chen2022regionvit}
Chun-Fu Chen, Rameswar Panda, and Quanfu Fan.
\newblock {RegionViT}: Regional-to-local attention for vision transformers.
\newblock In \emph{International Conference on Learning Representations (ICLR)}, 2022.

\bibitem[Chen et~al.(2023)Chen, Liu, Tang, Yi, Zhao, and Han]{chen2023sparsevit}
Xuanyao Chen, Zhijian Liu, Haotian Tang, Li~Yi, Hang Zhao, and Song Han.
\newblock {SparseViT}: Revisiting activation sparsity for efficient high-resolution vision transformer.
\newblock In \emph{IEEE/CVF Conference on Computer Vision and Pattern Recognition (CVPR)}, 2023.

\bibitem[Cordonnier et~al.(2021)Cordonnier, Mahendran, Dosovitskiy, Weissenborn, Uszkoreit, and Unterthiner]{cordonnier2021dps}
Jean-Baptiste Cordonnier, Aravindh Mahendran, Alexey Dosovitskiy, Dirk Weissenborn, Jakob Uszkoreit, and Thomas Unterthiner.
\newblock Differentiable patch selection for image recognition.
\newblock In \emph{IEEE/CVF Conference on Computer Vision and Pattern Recognition (CVPR)}, 2021.

\bibitem[Cuturi(2013)]{cuturi2013sinkhorn}
Marco Cuturi.
\newblock Sinkhorn distances: Lightspeed computation of optimal transport.
\newblock In \emph{Advances in Neural Information Processing Systems}, 2013.

\bibitem[Dao et~al.(2022)Dao, Fu, Ermon, Rudra, and R{\'e}]{dao2022flashattention}
Tri Dao, Daniel~Y. Fu, Stefano Ermon, Atri Rudra, and Christopher R{\'e}.
\newblock {FlashAttention}: Fast and memory-efficient exact attention with {IO}-awareness.
\newblock In \emph{Advances in Neural Information Processing Systems}, 2022.

\bibitem[Dosovitskiy et~al.(2021)Dosovitskiy, Beyer, Kolesnikov, Weissenborn, Zhai, Unterthiner, Dehghani, Minderer, Heigold, Gelly, Uszkoreit, and Houlsby]{dosovitskiy2021vit}
Alexey Dosovitskiy, Lucas Beyer, Alexander Kolesnikov, Dirk Weissenborn, Xiaohua Zhai, Thomas Unterthiner, Mostafa Dehghani, Matthias Minderer, Georg Heigold, Sylvain Gelly, Jakob Uszkoreit, and Neil Houlsby.
\newblock An image is worth 16x16 words: Transformers for image recognition at scale.
\newblock In \emph{International Conference on Learning Representations (ICLR)}, 2021.

\bibitem[Fayyaz et~al.(2022)Fayyaz, Koohpayegani, Jafari, Sengupta, Joze, Sommerlade, Pirsiavash, and Gall]{fayyaz2022ats}
Mohsen Fayyaz, Soroush~Abbasi Koohpayegani, Farnoush~Rezaei Jafari, Sunando Sengupta, Hamid Reza~Vaezi Joze, Eric Sommerlade, Hamed Pirsiavash, and J{\"u}rgen Gall.
\newblock Adaptive token sampling for efficient vision transformers.
\newblock In \emph{European Conference on Computer Vision (ECCV)}, LNCS, pp.\  396--414, 2022.

\bibitem[Grainger et~al.(2023)Grainger, Paniagua, Song, Cuntoor, Lee, and Wu]{grainger2023paca}
Ryan Grainger, Thomas Paniagua, Xi~Song, Naresh Cuntoor, Mun~Wai Lee, and Tianfu Wu.
\newblock {PaCa-ViT}: Learning patch-to-cluster attention in vision transformers.
\newblock In \emph{IEEE/CVF Conference on Computer Vision and Pattern Recognition (CVPR)}, 2023.

\bibitem[He et~al.(2016)He, Zhang, Ren, and Sun]{he2016resnet}
Kaiming He, Xiangyu Zhang, Shaoqing Ren, and Jian Sun.
\newblock Deep residual learning for image recognition.
\newblock In \emph{Proceedings of the IEEE Conference on Computer Vision and Pattern Recognition (CVPR)}, pp.\  770--778, 2016.

\bibitem[Huang et~al.(2023)Huang, Zhou, Cao, He, and Tan]{huang2023supertoken}
Huaibo Huang, Xiaoqiang Zhou, Jie Cao, Ran He, and Tieniu Tan.
\newblock Vision transformer with super token sampling.
\newblock In \emph{IEEE/CVF Conference on Computer Vision and Pattern Recognition (CVPR)}, 2023.

\bibitem[Jaderberg et~al.(2015)Jaderberg, Simonyan, Zisserman, and Kavukcuoglu]{jaderberg2015stn}
Max Jaderberg, Karen Simonyan, Andrew Zisserman, and Koray Kavukcuoglu.
\newblock Spatial transformer networks.
\newblock In \emph{Advances in Neural Information Processing Systems (NeurIPS)}, 2015.

\bibitem[Jaegle et~al.(2021)Jaegle, Gimeno, Brock, Zisserman, Vinyals, and Carreira]{jaegle2021perceiver}
Andrew Jaegle, Felix Gimeno, Andy Brock, Andrew Zisserman, Oriol Vinyals, and Joao Carreira.
\newblock Perceiver: General perception with iterative attention.
\newblock In \emph{International Conference on Machine Learning (ICML)}, 2021.

\bibitem[Jang et~al.(2017)Jang, Gu, and Poole]{jang2017gumbel}
Eric Jang, Shixiang Gu, and Ben Poole.
\newblock Categorical reparameterization with {Gumbel}-softmax.
\newblock In \emph{International Conference on Learning Representations}, 2017.

\bibitem[Jiang et~al.(2026)Jiang, Zhu, Huang, Zhang, and Li]{jiang2026lens}
Kai Jiang, Ruishu Zhu, Siqi Huang, Hongyuan Zhang, and Xuelong Li.
\newblock Latent noise mask for reducing visual redundancy in multimodal large language models.
\newblock arXiv:2606.30168, 2026.

\bibitem[Jiang et~al.(2024)Jiang, Zhang, Wang, Zhang, Wang, and Du]{jiang2024lemevit}
Wentao Jiang, Jing Zhang, Di~Wang, Qiming Zhang, Zengmao Wang, and Bo~Du.
\newblock {LeMeViT}: Efficient vision transformer with learnable meta tokens for remote sensing image interpretation.
\newblock In \emph{International Joint Conference on Artificial Intelligence (IJCAI)}, 2024.

\bibitem[Katharopoulos \& Fleuret(2019)Katharopoulos and Fleuret]{katharopoulos2019attsampling}
Angelos Katharopoulos and Fran{\c{c}}ois Fleuret.
\newblock Processing megapixel images with deep attention-sampling models.
\newblock In \emph{International Conference on Machine Learning (ICML)}, 2019.

\bibitem[Krizhevsky(2009)]{krizhevsky2009cifar}
Alex Krizhevsky.
\newblock Learning multiple layers of features from tiny images.
\newblock Technical report, University of Toronto, 2009.

\bibitem[Le \& Yang(2015)Le and Yang]{le2015tinyimagenet}
Ya~Le and Xuan Yang.
\newblock Tiny {ImageNet} visual recognition challenge.
\newblock Technical report, Stanford University, CS231N course report, 2015.

\bibitem[LeCun et~al.(1998)LeCun, Bottou, Bengio, and Haffner]{lecun1998mnist}
Yann LeCun, L{\'e}on Bottou, Yoshua Bengio, and Patrick Haffner.
\newblock Gradient-based learning applied to document recognition.
\newblock \emph{Proceedings of the IEEE}, 86\penalty0 (11):\penalty0 2278--2324, 1998.

\bibitem[Lee et~al.(2019)Lee, Lee, Kim, Kosiorek, Choi, and Teh]{lee2019settransformer}
Juho Lee, Yoonho Lee, Jungtaek Kim, Adam Kosiorek, Seungjin Choi, and Yee~Whye Teh.
\newblock Set transformer: A framework for attention-based permutation-invariant neural networks.
\newblock In \emph{International Conference on Machine Learning (ICML)}, 2019.

\bibitem[Li et~al.(2026)Li, Wang, Liu, Li, and Guo]{representative2026}
Yuntong Li, Hainuo Wang, Hengxing Liu, Mingjia Li, and Xiaojie Guo.
\newblock Representative attention for vision transformers.
\newblock arXiv:2605.14913, 2026.

\bibitem[Liu et~al.(2021)Liu, Lin, Cao, Hu, Wei, Zhang, Lin, and Guo]{liu2021swin}
Ze~Liu, Yutong Lin, Yue Cao, Han Hu, Yixuan Wei, Zheng Zhang, Stephen Lin, and Baining Guo.
\newblock Swin transformer: Hierarchical vision transformer using shifted windows.
\newblock In \emph{IEEE/CVF International Conference on Computer Vision (ICCV)}, 2021.

\bibitem[Locatello et~al.(2020)Locatello, Weissenborn, Unterthiner, Mahendran, Heigold, Uszkoreit, Dosovitskiy, and Kipf]{locatello2020slot}
Francesco Locatello, Dirk Weissenborn, Thomas Unterthiner, Aravindh Mahendran, Georg Heigold, Jakob Uszkoreit, Alexey Dosovitskiy, and Thomas Kipf.
\newblock Object-centric learning with slot attention.
\newblock In \emph{Advances in Neural Information Processing Systems (NeurIPS)}, 2020.

\bibitem[Mnih et~al.(2014)Mnih, Heess, Graves, and Kavukcuoglu]{mnih2014ram}
Volodymyr Mnih, Nicolas Heess, Alex Graves, and Koray Kavukcuoglu.
\newblock Recurrent models of visual attention.
\newblock In \emph{Advances in Neural Information Processing Systems (NeurIPS)}, 2014.

\bibitem[Rao et~al.(2021)Rao, Zhao, Liu, Lu, Zhou, and Hsieh]{rao2021dynamicvit}
Yongming Rao, Wenliang Zhao, Benlin Liu, Jiwen Lu, Jie Zhou, and Cho-Jui Hsieh.
\newblock {DynamicViT}: Efficient vision transformers with dynamic token sparsification.
\newblock In \emph{Advances in Neural Information Processing Systems (NeurIPS)}, 2021.

\bibitem[Raposo et~al.(2024)Raposo, Ritter, Richards, Lillicrap, Humphreys, and Santoro]{raposo2024mod}
David Raposo, Sam Ritter, Blake Richards, Timothy Lillicrap, Peter~Conway Humphreys, and Adam Santoro.
\newblock Mixture-of-depths: Dynamically allocating compute in transformer-based language models.
\newblock arXiv:2404.02258, 2024.

\bibitem[Ryoo et~al.(2021)Ryoo, Piergiovanni, Arnab, Dehghani, and Angelova]{ryoo2021tokenlearner}
Michael~S. Ryoo, AJ~Piergiovanni, Anurag Arnab, Mostafa Dehghani, and Anelia Angelova.
\newblock {TokenLearner}: Adaptive space-time tokenization for videos.
\newblock In \emph{Advances in Neural Information Processing Systems (NeurIPS)}, 2021.

\bibitem[Shan et~al.(2025)Shan, Wang, Zhao, Cai, Zhang, and Liritzis]{anchorformer2025}
Jiquan Shan, Junxiao Wang, Lifeng Zhao, Liang Cai, Hongyuan Zhang, and Ioannis Liritzis.
\newblock {AnchorFormer}: Differentiable anchor attention for efficient vision transformer.
\newblock arXiv:2505.16463, 2025.

\bibitem[Song et~al.(2026)Song, Chen, Nan, Ramanan, Tarr, and Luo]{song2026veca}
Alan~Z. Song, Yinjie Chen, Mu~Nan, Deva Ramanan, Michael~J. Tarr, and Andrew~F. Luo.
\newblock Do vision transformers need all-to-all attention? global communication through elastic learned cores.
\newblock arXiv:2605.12491, 2026.

\bibitem[Tabernik \& Sko{\v{c}}aj(2020)Tabernik and Sko{\v{c}}aj]{tabernik2020dfg}
Domen Tabernik and Danijel Sko{\v{c}}aj.
\newblock Deep learning for large-scale traffic-sign detection and recognition.
\newblock \emph{IEEE Transactions on Intelligent Transportation Systems}, 21\penalty0 (4):\penalty0 1427--1440, 2020.

\bibitem[Touvron et~al.(2021)Touvron, Cord, Douze, Massa, Sablayrolles, and J{\'e}gou]{touvron2021deit}
Hugo Touvron, Matthieu Cord, Matthijs Douze, Francisco Massa, Alexandre Sablayrolles, and Herv{\'e} J{\'e}gou.
\newblock Training data-efficient image transformers \& distillation through attention.
\newblock In \emph{International Conference on Machine Learning (ICML)}, 2021.

\bibitem[Wang et~al.(2022)Wang, Xie, Li, Fan, Song, Liang, Lu, Luo, and Shao]{wang2022pvtv2}
Wenhai Wang, Enze Xie, Xiang Li, Deng-Ping Fan, Kaitao Song, Ding Liang, Tong Lu, Ping Luo, and Ling Shao.
\newblock {PVT v2}: Improved baselines with pyramid vision transformer.
\newblock \emph{Computational Visual Media}, 8\penalty0 (3):\penalty0 415--424, 2022.

\bibitem[Xia et~al.(2022)Xia, Pan, Song, Li, and Huang]{xia2022dat}
Zhuofan Xia, Xuran Pan, Shiji Song, Li~Erran Li, and Gao Huang.
\newblock Vision transformer with deformable attention.
\newblock In \emph{IEEE/CVF Conference on Computer Vision and Pattern Recognition (CVPR)}, 2022.

\bibitem[Yang et~al.(2023)Yang, Xu, De~Mello, Crowley, and Wang]{yang2023gpvit}
Chenhongyi Yang, Jiarui Xu, Shalini De~Mello, Elliot~J. Crowley, and Xiaolong Wang.
\newblock {GPViT}: A high resolution non-hierarchical vision transformer with group propagation.
\newblock In \emph{International Conference on Learning Representations}, 2023.

\bibitem[Yin et~al.(2022)Yin, Vahdat, Alvarez, Mallya, Kautz, and Molchanov]{yin2022avit}
Hongxu Yin, Arash Vahdat, Jose~M. Alvarez, Arun Mallya, Jan Kautz, and Pavlo Molchanov.
\newblock {A-ViT}: Adaptive tokens for efficient vision transformer.
\newblock In \emph{Proceedings of the IEEE/CVF Conference on Computer Vision and Pattern Recognition (CVPR)}, pp.\  10799--10808, 2022.

\bibitem[Yue et~al.(2021)Yue, Sun, Kuang, Wei, Torr, Zhang, and Lin]{yue2021psvit}
Xiaoyu Yue, Shuyang Sun, Zhanghui Kuang, Meng Wei, Philip Torr, Wayne Zhang, and Dahua Lin.
\newblock Vision transformer with progressive sampling.
\newblock In \emph{IEEE/CVF International Conference on Computer Vision (ICCV)}, 2021.

\bibitem[Zeng et~al.(2022)Zeng, Jin, Liu, Qian, Luo, Ouyang, and Wang]{zeng2022tcformer}
Wang Zeng, Sheng Jin, Wentao Liu, Chen Qian, Ping Luo, Wanli Ouyang, and Xiaogang Wang.
\newblock Not all tokens are equal: Human-centric visual analysis via token clustering transformer.
\newblock In \emph{IEEE/CVF Conference on Computer Vision and Pattern Recognition (CVPR)}, 2022.

\bibitem[Zeng et~al.(2024)Zeng, Jin, Xu, Liu, Qian, Ouyang, Luo, and Wang]{zeng2024tcformer}
Wang Zeng, Sheng Jin, Lumin Xu, Wentao Liu, Chen Qian, Wanli Ouyang, Ping Luo, and Xiaogang Wang.
\newblock {TCFormer}: Visual recognition via token clustering transformer.
\newblock \emph{IEEE Transactions on Pattern Analysis and Machine Intelligence}, 46\penalty0 (12):\penalty0 9521--9535, 2024.

\bibitem[Zhu et~al.(2023)Zhu, Wang, Ke, Zhang, and Lau]{zhu2023biformer}
Lei Zhu, Xinjiang Wang, Zhanghan Ke, Wayne Zhang, and Rynson W.~H. Lau.
\newblock {BiFormer}: Vision transformer with bi-level routing attention.
\newblock In \emph{IEEE/CVF Conference on Computer Vision and Pattern Recognition (CVPR)}, pp.\  10323--10333, 2023.

\end{thebibliography}
\bibliographystyle{tmlr}

\appendix

\section{The pilot model}
\label{app:pilot}
The pilot (RART-32) is the block of Fig.~\ref{fig:block} with \textsc{perceiver} routing (32
learned anchor queries, softmax over patches, softmax read-back), embedding width 384, anchor width
128, depth 6, on CIFAR-100 at $224^2$ ($14\times14$ patches), trained for 10 epochs with AdamW
(learning rate $3\cdot10^{-4}$, cosine schedule), one seed. It reached 39.73\% test accuracy,
against 45.44\% for a ViT of the same width and depth and 42.21\% for the same model with 16
anchors. The anchors' pairwise cosine similarity in blocks 1--6 was 0.9887, 0.9865, 0.9900,
0.9920, 0.9935 and 0.9950, the mean attention entropy over patches 5.272 (maximum 5.278 for 196
patches), and the mean similarity between anchor attention maps 0.9995.

\section{Training recipes}
\label{app:recipes}
\begin{table}[h]
\centering
\caption{Training settings. All runs: AdamW, weight decay 0.05, cosine schedule, from scratch.
``Clip'': gradient-norm clipping; ``warmup'': linear, in epochs.}
\begin{tabular}{llllll}
\toprule
& Collapse & Cluttered, 10 ep. & Cluttered, 30 ep. & Tiny IN & High resolution \\
\midrule
Data & CIFAR-100 & CIFAR-10 canvas & CIFAR-10 canvas & Tiny IN canvas & MMNIST / DFG \\
Epochs & 10 & 10 & 30 & 15 & 30 / 20 \\
Batch & 128 & 128 & 128 & 128 & 32 \\
Learning rate & $3\cdot10^{-4}$ & $3\cdot10^{-4}$ & $1.25\cdot10^{-4}$ & $1.25\cdot10^{-4}$ & $2\cdot10^{-4}$ \\
Clip / warmup & -- & --$^{\ast}$ & 1.0 / 2 & 1.0 / 2 & 1.0 / 2 \\
Augmentation & flip & flip & flip & flip & none \\
Precision & fp16 & fp16 / bf16 & bf16 & bf16 & bf16 \\
\bottomrule
\end{tabular}

\smallskip
{\raggedright\footnotesize $^{\ast}$Clip 1.0 / warmup 2 for the variants of Table~\ref{tab:ablation} and the
skipping runs marked ``clip''. ResNet-18: learning rate $10^{-3}$ for the 30-epoch, Tiny ImageNet
and high-resolution runs.\par}
\end{table}

\section{Reimplementing AnchorFormer and Representative Attention}
\label{app:reimpl}
Neither method has released code (the Representative Attention repository contained only a README
and figures in November 2026). \textbf{AnchorFormer}~\citep{anchorformer2025}: keys and values are
projected per head; $m$ learned anchors per head score every key, $A_{ij}=\mathrm{softmax}_j(w_j
k_i^{\top}/\sqrt{d})$; with $\Delta=\mathrm{diag}(A^{\top}\mathbf{1})$ the output is
$H=A\Delta^{-1}A^{\top}V$ (their Eqs.~11--13), heads concatenated and projected. The mass
normalisation $\Delta^{-1}$ is the only stabiliser the paper describes. \textbf{Representative
Attention}~\citep{representative2026}: per-head learned representatives $W_g$; assignments
$\mathrm{softmax}_m(KW_g)$, unscaled as in their appendix Eq.~14; slot-mass normalisation
$\hat A_{n,m}=A_{n,m}/(\sum_t A_{t,m}+\epsilon)$, $\epsilon=10^{-6}$; latent keys and values
$\hat A^{\top}K$, $\hat A^{\top}V$, layer-normalised; residual self-attention among the latents
with queries, keys and values derived from the normalised latent values; patches read the refined
latents by query attention over the normalised latent keys; and a $3\times3$ depth-wise
convolution on the patch values is added before the output projection (their Eqs.~2--10, kernel
size 3). Both modules use width 128 and 4 heads like the other pathways (10.1M and 10.5M
parameters in total, against 11.3M for the partition) and replace only the anchor pathway of our
block; the published models use them as the only attention.

\section{Per-seed results}
\label{app:perseed}
\begin{table}[h]
\centering
\caption{Paired gaps \focus{} $-$ \fixed{} on Cluttered CIFAR-10 at 10 epochs.}
\begin{tabular}{lccccc}
\toprule
& seed 42 & seed 43 & seed 44 & mean & 95\% CI \\
\midrule
4 anchors & +1.71 & +1.23 & +1.51 & +1.48 & [+0.88, +2.08] \\
16 anchors & +1.53 & +2.90 & +4.96 & +3.13 & [$-$1.16, +7.42] \\
\bottomrule
\end{tabular}
\end{table}
Individual seeds of every run are listed in the repository's result files and summary.

\section{Placement sensitivity}
\label{app:sensitivity}
\begin{table}[h]
\centering
\caption{\focus-16 on Cluttered CIFAR-10, 10 epochs, clip and warmup, seed 42; one parameter
changed at a time from $\beta=10$, $\sigma_s=1.5$ patches, $\sigma_h=0.35$ (reference 53.68).}
\begin{tabular}{lcc}
\toprule
Setting & Test accuracy (\%) & $\Delta$ \\
\midrule
$\beta=3$ & 53.98 & +0.30 \\
$\beta=30$ & 49.33 & $-$4.35 \\
$\sigma_s=0.75$ & 53.59 & $-$0.09 \\
$\sigma_s=3$ & 21.88 & $-$31.80 \\
$\sigma_h=0.15$ & 52.90 & $-$0.78 \\
$\sigma_h=1.0$ & 52.26 & $-$1.42 \\
\bottomrule
\end{tabular}
\end{table}
In the original 10-epoch recipe (no clipping), removing the home prior cost 2.6 points
(47.23 $\pm$ 1.17 against 49.84 $\pm$ 0.26), removing the coverage grid or using a $3\times3$ grid
made no measurable difference (49.63 and 49.40), and removing suppression diverged in one of
three seeds.

\section{Skipping per-patch computation}
\label{app:skipping}
From block 3 on, only a fraction of patches go through the FFN; the others skip it but have already
received their anchor's global update. Keeping 25\% of patches removes 25\% of the linear-layer
FLOPs (5.31 to 3.96 GFLOPs). With clipping and warmup, random selection costs $-$0.38 points
([$-$1.48, +0.71]; 53.40 $\pm$ 0.35 against 53.78 $\pm$ 0.09). Selecting by the learned importance
map is not better than random or token-norm selection (seed 42: 49.25 against 50.49 and 50.87),
and training through a straight-through importance gate is fragile: a ratio gate destabilised
training at 50\% kept (38.25\%), and a bounded gate $1+\pi-\mathrm{sg}(\pi)$ diverged in every
run, also with clipping.

\section{DFG pipeline checks}
\label{app:dfg}
Before trusting the near-chance DFG scores with average pooling (Table~\ref{tab:avgpool}) we
checked the pipeline: our mean average precision equals scikit-learn's on random data (57.850
both); rendered frames show the label boxes on the signs; the labels have 39--392 positive
training images per class; and an ImageNet-pretrained ResNet-18 memorises 64 training frames to
100 mAP in 30 passes. The same pretrained network fine-tuned on all frames for 6 epochs reaches
8.74 mAP with average pooling and 47.76 with max pooling.
\begin{table}[h]
\centering
\caption{DFG with average-pooled heads, seed 42, 20 epochs, test mAP (chance 3.1).}
\label{tab:avgpool}
\begin{tabular}{lcccccc}
\toprule
& \focus-64 & \fixed-64 & \densewin & Dense ViT & ViT 480-wide & ResNet-18 \\
\midrule
mAP & 8.10 & 8.14 & 12.35 & 11.02 & 8.76 & 8.33 \\
\bottomrule
\end{tabular}
\end{table}
A first Megapixel MNIST setting, $1024^2$ canvases with 50 noise patches and 5{,}000 fixed
training layouts, did not train: focus anchors stayed at chance (11.6\% after 10 epochs, training
loss at $\ln 10$). The $512^2$ setting with fresh layouts was chosen after a 3-epoch probe.

\section{Cost at $256^2$}
\label{app:cost}
\begin{table}[h]
\centering
\caption{Cost at $256^2$ on one A100 (bf16, untrained, measured alone), training step at batch
128.}
\begin{tabular}{lrrrr}
\toprule
Model & GFLOPs & Train step & Train memory & Inference \\
\midrule
\focus-16 & 5.49 & 99.3 ms & 4.53 GB & 5{,}259 img/s \\
\focus-4 & 5.45 & 70.8 ms & 4.24 GB & 6{,}207 img/s \\
\fixed-16 & 5.49 & 60.7 ms & 4.31 GB & 6{,}408 img/s \\
ViT, $256^2$ & 5.28 & 32.4 ms & 3.11 GB & 11{,}176 img/s \\
ViT, $128^2$ & 1.21 & 20.6 ms & 1.17 GB & 39{,}339 img/s \\
PVTv2-B0 & 1.51 & 52.4 ms & 4.94 GB & 6{,}310 img/s \\
\bottomrule
\end{tabular}
\end{table}

\end{document}